\documentclass[11pt]{article}

\usepackage[final]{acl}

\usepackage{times}
\usepackage{latexsym}

\usepackage[T1]{fontenc}

\usepackage[utf8]{inputenc}

\usepackage{microtype}

\usepackage{inconsolata}

\usepackage[multiple]{footmisc}

\usepackage{graphicx}

\usepackage{url}
\title{SAFARI: A Community-Engaged Approach and Dataset of Stereotype Resources in the Sub-Saharan African Context}
\date{August 2025}

\author{
    Aishwarya Verma \\ Google Research\\ \texttt{aishv@google.com} 
    \And
    Laud Ammah\thanks{\hspace{5pt}These authors contributed equally to this work.} \\ RAIN Africa \\ \texttt{laud.ammah@tum.de} 
    \And 
    Olivia Nercy Ndlovu Lucas\footnotemark[1]\, \thanks{\hspace{5pt}This work was completed while the author was a contractor at Mantaray Africa.} \\ Mantaray Africa\\\texttt{massireninercy.n@gmail.com}
    \AND
    Andrew Zaldivar \\ Google Research\\ \texttt{andrewzaldivar@google.com} 
    \And 
    Vinodkumar Prabhakaran \\ Google Research\\ \texttt{vinodkpg@google.com} 
    \And
    Sunipa Dev \\ Google Research\\ \texttt{sunipadev@google.com}
}
\usepackage{booktabs}
\usepackage{multirow}
\usepackage{comment}
\begin{document}

\newcommand{\sd}[1]{\textcolor{purple}{#1 -sd}}
\maketitle
\begin{abstract}
Stereotype repositories are critical to assess generative AI model safety, but currently lack adequate global coverage. It is imperative to prioritize targeted expansion, strategically addressing existing deficits, over merely increasing data volume. This work introduces a multilingual stereotype resource covering four sub-Saharan African countries that are severely underrepresented in NLP resources: Ghana, Kenya, Nigeria, and South Africa. By utilizing socioculturally-situated, community-engaged methods, including telephonic surveys moderated in native languages, we establish a  reproducible methodology that is sensitive to the region's complex linguistic diversity and traditional orality. By deliberately balancing the sample across diverse ethnic and demographic backgrounds, we ensure broad coverage, resulting in a dataset of 3,534 stereotypes in English and 3,206 stereotypes across 15 native languages. \textcolor{red}{\textit{Content Warning: This paper contains examples of stereotypes that may be offensive}.}
\end{abstract}

\section{Introduction}

With rapid advances in generative AI~\cite{achiam2023gpt, anil2023palm} and its widespread adoption~\cite{why_chatgpt_2025}, there is growing focus on ensuring its utility and safety across the globe ~\cite{Jernite_2022,parrish2023adversarialnibblerdatacentricchallenge}. Recent work has investigated model safety failures in different parts of the world~\cite{birhane2021multimodaldatasetsmisogynypornography, ganguli2022redteaminglanguagemodels, kirk-etal-2022-handling}, with a particular focus on compiling lists of stereotypes to assess how these models propagate stereotypes about communities worldwide~\cite{mitchell-etal-2025-shades, parrish-etal-2022-bbq}. These lists are sourced through different methods, including LLM-based generation~\cite{jha2023seegullstereotypebenchmarkbroad}, large-scale surveys~\cite{dev2023spice}, or annotation studies~\cite{bhutani-etal-2024-seegull-1}. 
However, significant gaps persist in global coverage, 
along with concerns about the data collection approaches themselves~\cite{smart2024sociallyresponsibledatalarge}.

\begin{figure*}[ht!]
    \centering
    \includegraphics[width=0.75\textwidth]{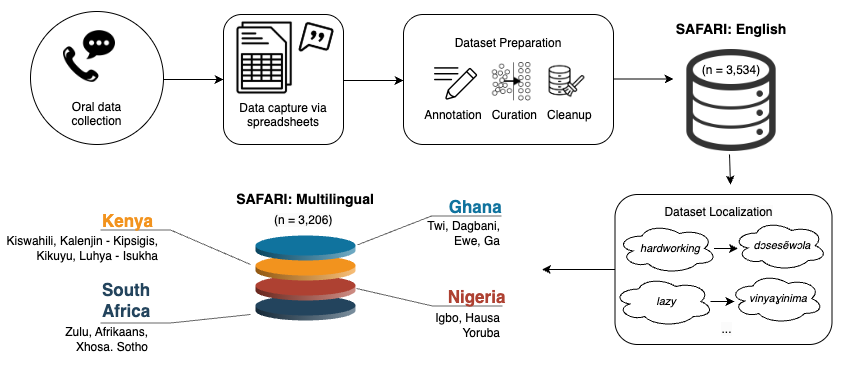}  
    \scriptsize
    \caption{Our community-engaged methodology towards curating the multilingual SAFARI dataset, comprising 3,206 stereotypes across four countries and 15 languages. See Table \ref{tab:multilingualstats} for the language distribution.}
    
    \label{fig:wide_diagram}
\end{figure*}

Nowhere are these gaps more evident than in sub-Saharan Africa, where NLP datasets have historically lacked adequate coverage and quality~\cite{kreutzer-etal-2022-quality-1, nekoto-etal-2020-participatory-1, ousidhoum-etal-2025-building}. While recent efforts such as the Masakhane project,\footnote{\url{https://www.masakhane.io/}} and Africa-focused research on tasks like machine translation~\cite{adelani-etal-2022-thousand-1}, hate speech detection~\cite{muhammad-etal-2025-afrihate}, cultural understanding \cite{ahmad2024generativelanguagemodelsmulticultural}, and language understanding in general~\cite{ojo-etal-2025-afrobench} have made significant leaps, there remains a lack of stereotype resources that capture the local cultural knowledge from the region. 

Overcoming this deficiency effectively requires a participatory and tailored approach to adapt large language models (LLMs) to the nuanced sociocultural realities of the continent~\cite{adebara-abdul-mageed-2022-towards, DeWittPrat2024DecolonizingLLMs, rashid2025amplifyinitiativebuildinglocalized,Oluka2024Mitigating}. Crucially, a truly participatory framework must also adapt the conventional Western data collection methods, as these are often incompatible with local contexts due to technological barriers like limited internet access, costly mobile data, and poor keyboard support for local languages~\cite{ade2023artificial,travaly2020future}.

To address these critical gaps in both data and methodology, this paper introduces the sub-\textbf{S}aharan \textbf{AF}rican \textbf{A}ssociations about \textbf{R}egionally-salient \textbf{I}dentities (SAFARI) dataset,\footnote{\url{https://github.com/google-research-datasets/SAFARI}} developed through an in-depth community-engaged process (see Figure~\ref{fig:wide_diagram}).

SAFARI utilizes telephonic surveys, conducted by local partners, to honor the region's rich oral traditions~\cite{Tchindjang2008Languages}---a direct counterpoint to prevailing traditional data collection paradigms in the West. Spanning four major sub-Saharan African countries with high populations (from the top-15)---Ghana, Kenya, Nigeria, and South Africa---the dataset contains stereotypes in 15 languages (in addition to English), each enriched with crucial local context such as perceived offensiveness and regional prevalence. We present SAFARI not only as a vital resource for building more equitable language models, but also as a methodological blueprint for more participatory data collection in AI. 

\section{Data Collection Methodology}

Recent work in the space of LLM Safety has demonstrated the importance of growing stereotype resources in community-engaged ways~\cite{davani2025comprehensiveframeworkoperationalizesocial}. These approaches introduce into consideration a vast majority of sensitive identities and prevalent stereotypes that are otherwise not safeguarded against, leading to major safety and representational failures in generative AI.  
 Our work builds upon some of these notable community- and people-centered approaches such as HESEIA in Latin America by ~\citet{ivetta2025heseia}, SPICE in India by \citet{dev2023spice}, and SHADES by ~\citet{mitchell-etal-2025-shades} globally. However, we find that these approaches fall short in the highly diverse, multiethnic, multilingual, as well as extremely underserved context of our target African countries -- where oral expression captures a lot more value and nuance. 
 We create a methodology that specifically caters to the languages and countries of interest.

\subsection{Survey Mode}
Given the linguistic complexity of African languages, which includes rich morphology and sophisticated tonal variations \cite{imam2025automaticspeechrecognitionafrican, Jerro2018LinguisticCA}, relying on written surveys~\cite{dev2023spice} or interactive written tools~\cite{ivetta2025heseia}, risked missing crucial nuances in oral communication. Our methodology, therefore, is grounded in a dual linguistic strategy \cite{Oyekunle2025Exploring}, wherein community knowledge is first gathered orally in the indigenous language and then recorded bilingually in the dataset, using both English and the native language. We leverage telephonic surveys with localized moderation in the respondent's preferred language, which allows a balance between broad reach and the ability to capture nuanced, authentic, and orally-expressed data.

\subsection{Local Partnerships}
This research study was conducted in collaboration with two organizations located in and operating within the countries of focus: Mantaray and RAIN Africa.\footnote{\url{https://mantaray.africa/}}\textsuperscript{,}\footnote{\url{https://rainafrica.org/}}
The recruitment of participants across communities was a collaborative effort between both of these organizations, with Mantaray also serving as the moderator for telephonic surveys in an array of languages commonly used in the countries of interest, and also preparing and translating the data.
 These partnerships were vital to ensure that local perspectives drive the salient questions of participant selection, demographic distribution, data quality, localization of survey questions, and more.
 
Our study recruited 410 participants, at least 100 from each country. We employed a controlled quota sampling strategy based on the distribution of major ethnic groups within each country. We initiated the study with a screener to record demographic data and ensure adherence to the desired participant distribution.  

\subsection{Interview Protocol}
We obtained informed consent from all participants and disclosed the monetary compensation amount (See details in Appendix \ref{app: data}) beforehand.  During the telephonic interview, participants were asked to share 10 to 15 prevalent stereotypes from their country (with a limit of three stereotypes per identity term to encourage diversity). They were also asked about the degree of offensiveness of the stereotype, its prevalence, and the axis of identity that the stereotype is about. 
These additional features will help identify the impact model failures can have when they perpetuate certain stereotypes. The survey in its entirety is shared in Appendix \ref{app: interview design}.

\subsection{Data Capture \& Localization}

Moderators were selected based on their fluency in the native languages as reported by the participants in the screener, 
in order to conduct interviews in each participant's preferred language. 
Despite this, we had to adopt a two-step approach that first captured the stereotypes in English which were then translated back to local languages by professional linguists. This was necessary because direct real-time transcription was often impractical (except for Swahili) due to linguistic and technical constraints. Moderators, while fluent speakers, often lacked the necessary written proficiency in native languages, a situation rooted in well-documented systemic challenges \cite{Appiah2021dilemma, Bamgbose2000language, Matavire2024Development, Mncwango2017diminished}. Furthermore,  standard electronic keyboards lack the specialized symbols and characters required for many languages from sub-Saharan Africa \cite{Chimkono2021UsabilityStudy}, and the unique phonetic signs inherent in these languages' orthographies cannot be represented by Roman characters. 

Except for Swahili in Kenya, the moderators initially recorded all responses in written English, where applicable. To recover the original linguistic nuance, professional linguists were subsequently hired to translate these standardized English responses back into the participant's native language, or to translate Swahili entries to English for completeness of our English dataset.

\paragraph{Notes on Retention of Nuance}
While this multi-step localization process introduces some complexity and potential mediation (as noted under Limitations), it was a deliberate trade-off made to maximize language coverage across a diverse set of languages (Table \ref{tab:multilingualstats}), given systemic challenges and resource constraints. To mitigate the inevitable loss of nuance during this process, we prioritized interview moderation in participants' native languages and the involvement of professional linguists to ensure that the final dataset remained as faithful as possible to the original intent of the participants.

We offer this work as a foundational step and encourage future research to develop methods that can further minimize mediation and expand authentic, high-fidelity multilingual stereotype resources representing under-resourced African languages.

\subsection{Data Preparation \& Annotation}
Following the initial data capture, the recorded stereotype statements were broken down and annotated into three key components as outlined in Table \ref{tab:glossary}. A single identity term can relate to multiple axes. For instance, the term `Hausa women' involves both the gender and ethnicity axes. Furthermore, identity axes can overlap. For example, the term `Northerners' may indicate both a regional identity and a cluster of specific ethnicities associated with the Northern region of Ghana.
Participants also provide additional social context about the stereotypes they share. They do so by rating each stereotype they shared on two perceived parameters: \textit{Offensiveness}, rated on a 5-point scale (1 = Not Offensive; 5 = Extremely Offensive), and \textit{Prevalence}, rated on a 4-point scale (1 = Rarely Used; 4 = Extremely Prevalent) (Appendix \ref{app:rating}).
Our choice to use self-reported metrics for offensiveness and prevalence was an intentional effort to center the participant’s voice, ensuring that the resulting dataset reflects a more granular and authentic range of perceptions regarding the stereotypes shared. We note that there will be inherent subjectivity in these ratings, which stem from individual calibrations of the assessment scales. 
\begin{table}[h]
\centering

\small
\begin{tabular}{p{0.15\columnwidth} p{0.46\columnwidth}  p{0.22\columnwidth}}
\toprule
\textbf{Name}& \textbf{Description}& \textbf{Example}\\
\midrule
Identity Term& A word or phrase used to describe a significant aspect of a personal or social identity.& Nigerians, Policemen,  Zulu Women\\
\midrule
Stereotype Attribute& The word associated with or used to describe the mentioned identity term. & Violent, Lazy, Ritualists, Liar\\
\midrule
Identity Axis& Broader categories that group related identity terms.& Ethnicity, Race, Gender\\
\bottomrule
\end{tabular}
\caption{Key terms used in the SAFARI dataset}
\label{tab:glossary}
\end{table}

\section{The SAFARI Dataset}

We introduce SAFARI (sub-\textbf{S}aharan \textbf{AF}rican \textbf{A}ssociations about \textbf{R}egionally-salient \textbf{I}dentities), a dataset comprising 3,534 stereotypes in English and 3,206 stereotypes in 15 native languages (listed in Table \ref{tab:multilingualstats}), collected from 410 participants across four countries. Specifically, it contains 1,919 stereotypes tied to ethnic groups in Ghana, Kenya, Nigeria and South Africa, and 382 stereotypes related to regions or states within these countries, thereby highlighting its granular, local focus. We describe the distribution of the stereotypes multilingually in Table \ref{tab:multilingualstats}.

\begin{table}[htb!]
\centering

\resizebox{\columnwidth}{!}{
\begin{tabular}{
    l 
    l 
    l 
    r 
}
\toprule
\textbf{Country} & \textbf{Language} & \textbf{ISO 639-3} & \textbf{Count} \\
\midrule
\multirow{5}{*}{\textbf{Kenya}}
    & Kiswahili & swa & 954 \\
    & Kalenjin (Kipsigis) & sgc & 157 \\
    & Kikuyu & kik & 132 \\
    & Luhya (Isukha) & ida & 115 \\
\cmidrule(lr){2-4}
    & \textbf{Kenya Total} & & \textbf{1358} \\
\midrule
\multirow{5}{*}{\textbf{Ghana}}
    & Twi & twi & 305 \\
    & Dagbani & dag & 211 \\
    & Ewe & ewe & 202 \\
    & Ga & gaa & 196 \\
\cmidrule(lr){2-4}
    & \textbf{Ghana Total} & & \textbf{914} \\
\midrule
\multirow{4}{*}{\textbf{Nigeria}}
    & Igbo & ibo & 184 \\
    & Hausa & hau & 170 \\
    & Yoruba & yor & 167 \\
\cmidrule(lr){2-4}
    & \textbf{Nigeria Total} & & \textbf{521} \\
\midrule
\multirow{5}{*}{\textbf{South Africa}}
    & Zulu & zul & 180 \\
    & Afrikaans & afr & 86 \\
    & Xhosa & xho & 75 \\
    & Sotho & sot & 72 \\
\cmidrule(lr){2-4}
    & \textbf{South Africa Total} & & \textbf{413} \\
\midrule
\multicolumn{3}{r}{\textbf{Grand Total}} & \textbf{3206} \\
\bottomrule
\end{tabular}
} 
\caption{Distribution of Multilingual Stereotypes by Country and Language}
\label{tab:multilingualstats}
\end{table}

\subsection{Dataset Characteristics}

\textbf{Diverse Participant Base}
The study cohort features a largely balanced gender distribution, with 52.4\% Male (n=215), 46.8\% Female (n=192), and 0.7\% Non-Binary (n=3) participants. Age representation in the cohort spanned a wide range, from individuals aged 18-24 up to the 65+ age group. We particularly aimed for diversity across ethnicity, tribes, and linguistic identity, as these elements form the core social fabric of our target countries. Our study cohort collectively represents 56 ethnic groups and speaks over 120 unique languages (with the majority of participants being at least bilingual).

\textbf{Overlap with Other Resources}
Our dataset has minimal overlap with common stereotype resources, all of which exhibit geographical limitations: StereoSet \cite{nadeem-etal-2021-stereoset-1}, SPICE \cite{dev2023spice} and HESEIA \cite{ivetta2025heseia} are limited to the perspectives from the USA, 
India, and Latin America, respectively. While SeeGULL Multilingual \cite{bhutani-etal-2024-seegull-1} covers Swahili (Kenya),  the Swahili subset of SAFARI only shares four common identity terms with it; in fact, we introduce 168 new, unique identity terms, not previously covered.
StereoSet does include seven African nationality-based identities, but the stereotypes were derived solely from US-based participants. In contrast, our resource features stereotypes shared by participants \textit{from} sub-Saharan Africa. 

\begin{figure}[t]
    \centering
    
    \includegraphics[width=0.95\columnwidth]{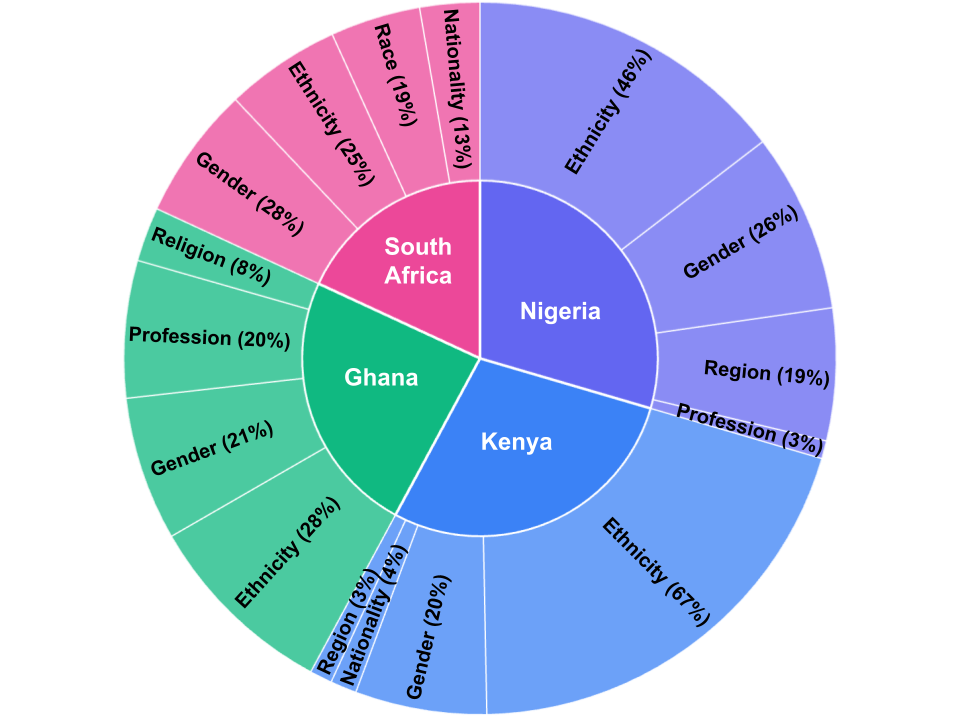}
    
    \caption{
     Country-wise distribution of dominant identity axes in the SAFARI stereotype dataset 
    }
    \label{fig:sunburst} 
\end{figure}

\textbf{Prominent Identity Axes:}
 While ethnicity and gender remain the most dominant identity axes of stereotypes across the entire dataset, country-specific distributions reveal unique patterns, as depicted in Figure \ref{fig:sunburst}. In Nigeria, region is the third-most dominant identity axis after ethnicity and gender. Ghana's focus shifts to profession as its third major axis. The pattern in South Africa is unique, with race and nationality placing highly due to the country's distinct racial diversity. 

\subsection{Analysis and Evaluations}
\begin{table*}[]
    \centering
    \small
    \begin{tabular}{l|cccccc}
    \toprule
        Country & Gemini 2.5 & GPT-4o & Claude Sonnet 4.5 & Gemini 3 Pro & GPT-5.1 & Claude Opus 4.5 \\ \midrule
        Kenya (Swahili)      & 23.6 (19.8) & 19.1 (16.8) & 21.8 (20.6) & 36.2 (38.8) & 22.0 (17.0) & 28.5 (26.9) \\
        Ghana (Akan)         & 9.2 (9.6)   & 4.5 (7.8)   & 12.9 (27.2) & 21.4 (34.4) & 9.3 (19.5) & 14.1 (29.5) \\
        Nigeria (Hausa)      & 8.8 (17.0)  & 8.1 (9.3)   & 11.0 (16.0) & 18.7 (30.7) & 7.8 (16.8) & 15.8 (30.2) \\
        South Africa (Zulu)  & 5.7 (5.2)   & 2.5 (3.2)   & 7.2 (8.3)   & 16.0 (18.6) & 4.5 (4.2)  & 12.1 (17.8) \\ \bottomrule        
    \end{tabular}
    \caption{Rate of stereotype endorsement by industry leading models.The rate of stereotyping in English language is stated outside the braces and in the language mentioned within the braces respectively.}
    \label{tab:evals}
\end{table*}
\paragraph{Common Stereotype Attributes:}
128 stereotypes mentioning `black magic,' `witchcraft,' `ritualistic', or `juju' are found in the dataset, with 82.8\% 
from Ghana and Kenya combined. These stereotypes primarily target ethnic groups, particularly the Ewe (Ghana), and the Kisii and Kamba (Kenya). 
152 stereotypes related to financial status or display of wealth (with mentions of `rich', `love money', `show off wealth', `extravagant') emerge. 61.8\% 
of these relate to ethnic groups, primarily associated with the Kikuyu (Kenya) and Igbo (Nigeria). 
 Of the 327 profession-related stereotypes in the dataset, 74.6\%  
emerge from Ghana,  and most of these carry a negative connotation, targeting professions like police (`corrupt', `take bribes'), politicians (`liars', `corrupt'), and nurses (`unprofessional', `disresepctful'). 

Out of the 1,034 gender-related stereotypes, 57.9\% 
target women, and the remaining 42.1\% 
target men.

\paragraph{Prevalence of Stereotypes in Generative AI:}
Stereotype datasets are used widely to both assess model skews and safety and to steer them towards fair answers. We thus test whether AI models endorse stereotypes from SAFARI in English or in languages from each of the four countries. We adopt the NLI based technique from ~\citet{Dev_Li_Phillips_Srikumar_2020} for this assessment. In this test, each data point consists of two sentences: a premise and a hypothesis each of which invoke the attribute term and identity term respectively. The NLI based test checks if the model infers one from the other without sufficient context, which confirms model's tendency to stereotype. We create and employ human translators to carefully translate the NLI-based test templates from ~\citet{Dev_Li_Phillips_Srikumar_2020} and evaluate the tendency to stereotype in generation~\cite{jha2023seegullstereotypebenchmarkbroad} by Gemini 2.5 and Gemini 3 Pro~\cite{comanici2025gemini25}, GPT-4o and GPT 5.1~\cite{openai2024gpt4technicalreport}, and Claude Sonnet 4.5 and Claude Opus 4.5~\cite{TheC3}. Table \ref{tab:evals} describes our findings across all six frontier models. We see how it underscores that not only do models stereotype in the respective commonly used languages of each country, but also in English. In fact the extent of stereotyping is higher in English in many cases. We hypothesize that this discrepancy between the regionally popular language and English is stemmed in how the models do not particularly support the regional language and often fail to generate meaningfully. However, the overall prevalence of stereotyping across the board highlights the need to account for these stereotypes from underrepresented regions of the world to prevent potentially harmful generations.

\section{Discussion and Future Work} 

We present the SAFARI dataset, that fills a critical data gap for AI safety in sub-Saharan Africa, while also contributing a novel methodology for culturally-situated data collection in NLP. Through a locally-situated and community-engaged approach, we collected over 3K stereotypes in English as well as 15 native languages from Ghana, Kenya, Nigeria, and South Africa. This dataset enables us to reveal previously undetected instances of stereotypes propagated by leading generative AI models, validating the urgent need for this work. 

Beyond research evaluation, the SAFARI dataset can be plugged seamlessly into the many ways in which other West-focused stereotype resources are currently used, including for bias mitigation of models via counterfactual data generation \cite{zmigrod-etal-2019-counterfactual-1} and few shot debiasing, content moderation \cite{davani-etal-2023-hate} and more. SAFARI can also be leveraged in approaches that semi automate stereotype curation based of few shot prompting \cite{bhutani-etal-2024-seegull-1}.

Furthermore, our use of moderated, in-language telephonic interviews represents a novel data collection approach for NLP. This method provides a practical blueprint for navigating the complex techno-linguistic challenges prevalent in many global regions, effectively bypassing obstacles like low digital access and languages with orthographies unsupported by standard keyboards.
Our work underscores the necessity of expanding data collection to more languages and global regions, while also accounting for such local realities, and we strongly encourage future research to build upon this foundation.

\newpage

\section*{Limitations}
Our work was aimed at improving the coverage of stereotype resources across languages prevalent in four countries in sub-Saharan Africa.  While we attempt to carefully increase this coverage alongside maintaining demographic diversity of participants, we were limited by both our methodology and resources available to us. In particular, telephonic interviews allowed us to orally gather and record stereotypes, and thus reduce barriers related to writing or typing, but we were still limited by connectivity issues.

In particular, we may have failed to reach participants in remote areas who lack phones or reliable phone connectivity. For collecting perspectives from these specific, harder-to-reach groups, in-person surveys may be a better-suited methodology.

Our localization approach, described in Section 2, required a multi-step translation process, that may risk the loss of some subtle linguistic nuance and context as presented in the original spoken form. This focus on language preservation is critical, particularly because moderators observed that a stereotype's offensiveness was often diluted when translated into English, suggesting a valuable future research avenue concerning the loss of semantic and emotional nuance during cross-lingual transfer.

Further, our resource, while foundational for studying stereotypes in the sub-Saharan African context, is not exhaustive. It covers the major ethnic groups and native languages within our target countries, but omits many other ethnicities and languages due to limited sample size and resource constraints. Future research must expand upon this foundation to achieve comprehensive coverage of the numerous additional countries, languages, and ethnicities in sub-Saharan Africa. We recommend cross-referencing the findings from this expanded work with established traditional stereotype literature from the region \cite{Fakunmoju2017Gender,HarneitSeivers2007IgboHA,Lawson2015Exploring,Naituli2018Examination,Tewolde2024Experiencing}.

\section*{Ethical Considerations}
Our research further highlights the high degree of linguistic complexity in sub-Saharan Africa. While our resource provides a significant subset of stereotypes towards ascertaining safety of generative models, they are in no way to be used to declare a model free of biases. For that, a lot more work and thought would be needed, including a much more exhaustive data collection. We also note that the recorded stereotypes should be only used for purposes of model evaluation and improvement as detailed in our Data Card (Appendix \ref{app: data}) in order to prevent misuse.

\section*{Acknowledgements}
First and foremost, we are extremely grateful to our data collection partners Mantaray and RAIN Africa, and their members and networks, whose expertise has been invaluable for this project. In particular, we thank Helga Stegmann, Liezel Stegmann, Michele Edwards, Partrick K. Wamuyu, Laeticia N. Onyejegbu, and Lavina Ramkissoon for sharing expert opinions. We thank all the moderators, data collection experts, project coordinators, and expert linguists who helped translate the stereotypes with socio-cultural grounding: Abiola Olaonipekun, Adisa Branice, Aisha Young, Alida Hillary, Ashleigh Burrell, Celeste Griesel, Darmack Kerubo, Deborah Agbeja, Dominic Abotchie, Emmanuel Mwadena, Feroza Khan, Geofrey Odhiambo, Gladys Allottey, Ilse Louw, John Ashihundu, Joyce Martins, Kennedy Mbithi, Kipkirui Ronald, Lindokuhle Modisane, Manoko Jegede, Margaret Muiruri, Martin Adjornor, Mary Obot, Mary Wangare, Mavis Dadzi, Naseega Adams, Nina Laubscher, Ogechi Anyanwu, Osumman Idrissu, Oyoe Quartey, Prince Danquah, Rahinatu Hamza, Richard Kueli, Samantha Lewis, Stephen Maina, Steve Amale, Sylvester Mukele, Victor Siele, and Yussif Abubakar.
We are also very grateful to Debojit Ghosh, Charu Kalia, Ding Wang, Roxane Ghinet, Kwaku Agbesi, Taylor Montgomery, Ronnelle Castro, Jamelle Louise Javillonar, Jamila Smith-Loud, and Saška Mojsilović for their partnership, generous feedback, and advice. Finally, we also thank the reviewers and ACs at ARR and EACL for their very constructive feedback and discussion.

\bibliography{custom,anthology,cultural,spice,references}

\begin{appendix}

\section*{Appendix}

\section{Dataset Details}
\label{app: data}
\subsection{Data}
The dataset is released at https://github.com/google-research-datasets/SAFARI, and is accompanied by
the data card, which specifies intended usage, data collection details, and more.

The dataset is structured into two tabs:
\begin{itemize}
    \item The first tab, 'safari\_english', contains 3,534 rows (excluding the header), each containing a stereotype and its corresponding annotations
    \item The second tab, 'safari\_multilingual', includes 3,206 rows (excluding the header), each containing a stereotype from the multilingual subset of the dataset,  covering 15 languages (following the distribution outlined in Table \ref{tab:multilingualstats}), along with corresponding annotations and the English counterpart for cross-reference.
\end{itemize}

The difference in the number of English (n=3,534) and multilingual (n=3,206) stereotypes is attributable to two factors: first, English-language interviews were not translated into a native language when English was the participant's preferred language; and second, in cases of less commonly spoken languages with limited linguist availability.

\subsection{Dataset Sample}
Below is a sample data entry to illustrate key parameters recorded in the dataset: 
\begin{itemize}
    \item \textbf{Country:} Kenya
    \item \textbf{Stereotype Sentence (English):} \textit{Mijikenda people are lazy}
    \item \textbf{Identity Term (English):} Mijikendas
    \item \textbf{Stereotype Attribute (English):} Lazy
    \item \textbf{Identity Axis (English):} Ethnicity
    \item \textbf{Offensiveness (1-5):} 3 - Slightly offensive
    \item \textbf{Prevalence (1-4):} 3 - Commonly used
    \item \textbf{Target Language:} Kiswahili
    \item \textbf{Identity Term (Kiwsahili):} Wamijikenda
    \item \textbf{Stereotype Attribute (Kiswahili):} Wavivu
\end{itemize}

A more detailed version is presented in the data card.

\subsection{Data Cleaning Steps}
Our data cleaning process was conservative and geared towards liberal inclusion of contributions from different participants. We focused solely on performing basic sanity checks to ensure the consistency needed for quantitative analyses, while deliberately avoiding extensive data tampering that could lead to further loss of contextual nuance collected from the field. No special software was used for this purpose. We cleaned the data manually and for aggregation purposes alone.

To clean the survey responses for precise counting and reporting, the following manual edits were applied:
\begin{itemize}
    \item Instances of "woman" and "women" were standardized to "women". Similarly, "Igbos" and "Igbo" (when referring to the people) were standardized to "Igbos", and other identity terms were treated similarly.
    \item Language names were standardized (e.g. 'Kiswahili' and 'Swahili' updated to 'Kiswahili'; 'isiZulu' and 'Zulu' updated to 'isiZulu', and so on).
    \item Other minor cleaning steps included correcting spelling errors and trimming extra white space within cells to prepare the data for usage and analysis.
\end{itemize}

\subsection{Review, Consent \& Compensation}
Dataset and data collection were reviewed by an internal, organizational, proprietary review board.

Prior to the study, informed consent was obtained from all participants, who confirmed they met the minimum age requirement (18+). The consent form explicitly stated their right to withdraw from the study at any time without penalty, and the option to decline any specific question asked to them. They were briefed on the study's purpose and the procedures for data retention, storage, and protection. In addition to the written consent form, the moderators ensured participants received these details verbally.

The dataset does not contain any Personally Identifiable Information (PII). All PII was scrubbed during collection. Unique, non-traceable identifiers (e.g., GH001, KE002, etc) were used solely for analysis and data counting purposes, but these cannot be used to trace or link back to the participating individuals.

For completing the telephonic survey, each participant was compensated in a fair and legally compliant manner with an amount that exceeded minimum local wage requirements. The incentive amount was mentioned in the screener shared beforehand. Duration of the telephonic interview was capped at 15 minutes per participant.

The two local organizations we partnered with were compensated either according to a pre-established Statement of Work (SOW) for services provided (in one case), or through a mutual agreement on the amount and deliverables (in the other). We will name both organizations and co-authors in the camera-ready version. 

\section{Interview Design and Methodology}
\label{app: interview design}

\subsection*{Section 1: Screener Demographic Questions}
The following information was collected during the initial screening phase:
\begin{enumerate}
    \item Country of Data Collection
    \item Gender Identity
    \item Age
    \item Hometown / Region
    \item Religion
    \item Ethnicity / Tribe
    \item Sub-Tribe (where applicable)
    \item Language(s) Spoken
    \begin{enumerate}
        \item Primary (Native)
        \item Secondary
        \item Other
    \end{enumerate}
    \item Highest Level of Completed Education
    \item Occupation
\end{enumerate}

\subsection*{Section 2: Stereotype Elicitation Questions}

Respondents were asked to share at most 10-15 stereotypes known in their society and community (as much as they could comfortably list within the limit of a 15-minute interview).

\begin{enumerate}
    \item \textbf{Question:} What stereotypes do you know in your society and community?
    \begin{enumerate}
        \item \textbf{Recording:} The stereotype was recorded, first as a sentence, and then broken down to an \emph{identity term} and an \emph{attribute term} (e.g., Nigerians, rich).
        \item \textbf{Axis of Identity:} The identity term's axis was classified (e.g., gender, tribe, religion, language, race, age, profession).
        \item \textbf{Offensiveness Rating:} Respondents were asked to rate the stereotype's offensiveness using a 1-5 scale (presented in Appendix \ref{app:rating}, Table \ref{tab:offensiveness_scale})
        \item \textbf{Prevalence Rating (Optional):} Respondents rated the stereotype's prevalence using a 1-4 scale (presented in Appendix \ref{app:rating}, Table \ref{tab:prevalence_scale})
           \end{enumerate}
    \item \textbf{English Translation:} The closest English translation was provided by the moderator post-interview, not during the collection process.
\end{enumerate}

\subsection*{Collection Guidelines}
To ensure diversity of responses, the following limits were applied per participant:
\begin{itemize}
    \item Not more than 3 stereotypes about any single identity term (e.g., Hausas).
    \item Not more than 5 stereotypes relating to any single identity axis (e.g., profession).
\end{itemize}

\section{Rating Scales}
\label{app:rating}
Participants were asked to rate the offensiveness of each shared stereotype on a 5-point scale (Table \ref{tab:offensiveness_scale}), and its prevalence on a 4-point scale (Table \ref{tab:prevalence_scale}).
\begin{table}
    \centering
    \begin{tabular}{cl}
        \toprule
        \textbf{Rating (Score)} & \textbf{Definition} \\
        \midrule
        1 & Not Offensive at all \\
        
        2 & Unsure \\
        
        3 & Slightly Offensive \\
        
        4 & Somewhat Offensive \\
        
        5 & Extremely Offensive \\
        \bottomrule
    \end{tabular}
    \caption{Offensiveness Rating Scale}
    \label{tab:offensiveness_scale}
\end{table}

\begin{table}[h!]
    \centering
    \begin{tabular}{cl}
        \toprule
        \textbf{Rating (Score)} & \textbf{Definition} \\
        \midrule
        1 & Rarely Used \\
        
        2 & Occasionally Used \\
        
        3 & Commonly Used \\
        
        4 & Extremely Prevalent/Popular \\
        \bottomrule
    \end{tabular}
    \caption{Prevalence Rating Scale}
    \label{tab:prevalence_scale}
\end{table}

\section{Data Details}
\label{app:datadetails}
\subsection{Participant Diversity}
The total number of stereotypes and participants per country is described in Table \ref{tab:countrystats}. The participants of our data collection effort were diverse along different demographic axes. Table \ref{tab:gender} describes the gender distribution of our participants per country.

\begin{table}[h!]
\centering

\resizebox{\columnwidth}{!}{
\begin{tabular}{l p{2.5cm} p{2.5cm}} 
\toprule
\textbf{Country} & \textbf{Number of Stereotypes} & \textbf{Number of Respondents} \\
\midrule
Ghana & 1037 & 101 \\
Kenya & 892 & 108 \\
Nigeria & 954 & 100 \\
South Africa & 651 & 101 \\
\midrule
\textbf{Total} & \textbf{3534} & \textbf{410} \\
\bottomrule
\end{tabular}
} 
\caption{Data Collection Summary by Country}
\label{tab:countrystats}
\end{table}

\begin{table}[h]
\centering

\begin{tabular}{l l r}
\toprule
\textbf{Country} & \textbf{Gender} & \textbf{Percentage (\%)} \\
\midrule
\multirow{2}{*}{Ghana} & Male & 54\% \\
& Female & 46\% \\
\midrule
\multirow{3}{*}{Kenya} & Female & 50\% \\
& Male & 49\% \\
& Non-binary & 1\% \\
\midrule
\multirow{2}{*}{Nigeria} & Male & 58\% \\
& Female & 42\% \\
\midrule
\multirow{3}{*}{South Africa} & Female & 50\% \\
& Male & 49\% \\
& Non-binary & 2\% \\
\bottomrule
\end{tabular}
\caption{Gender Distribution of Study Cohort by Country}
\label{tab:gender}
\end{table}

\subsection{Language Distribution}
\begin{figure}[h!]
    \centering
    \includegraphics[width=0.95\linewidth]{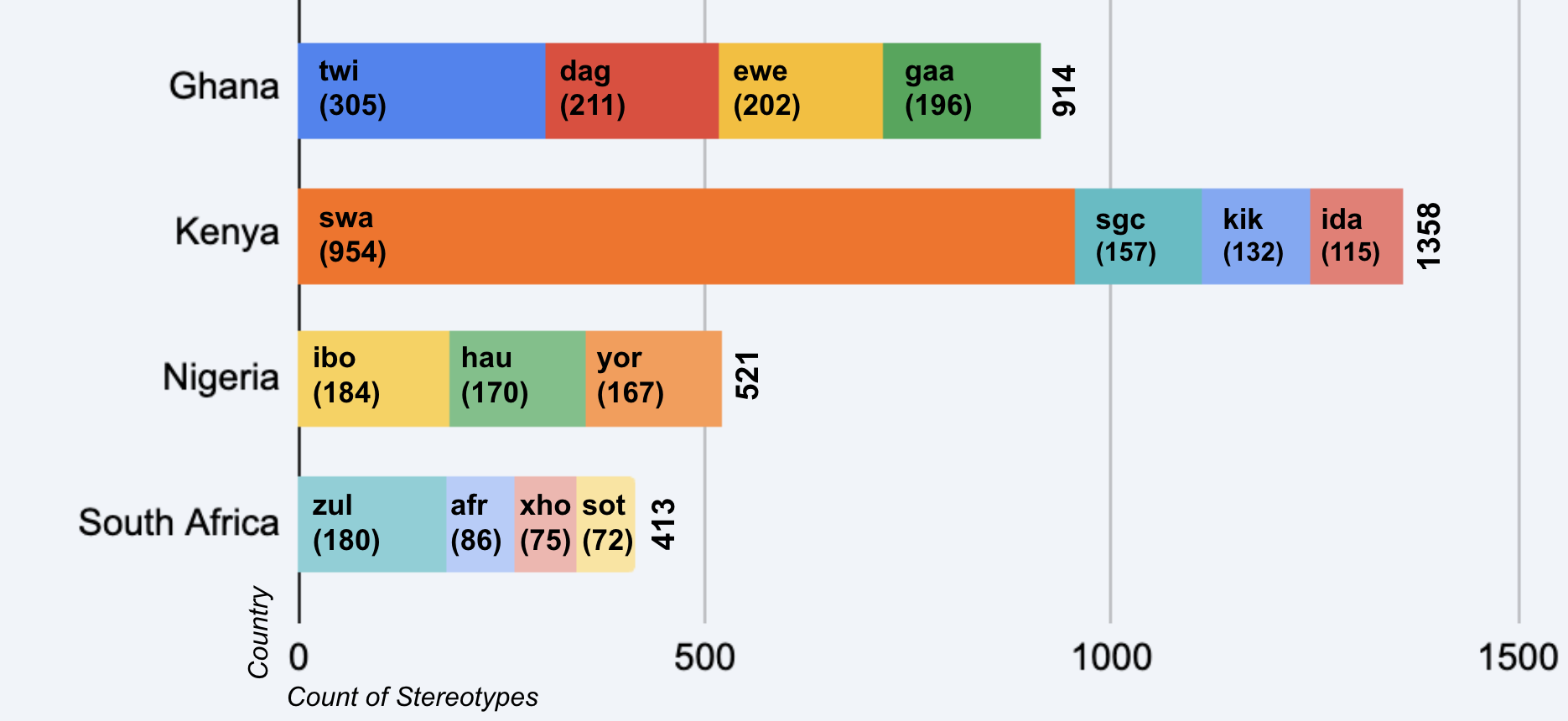} 
    \caption{The SAFARI dataset represents 15 native languages from 4 countries.  Table \ref{tab:multilingualstats} provides the expanded ISO codes.}
    \label{fig:langdistro}
\end{figure}

The language distribution of the multilingual stereotype dataset is shown in Figure \ref{fig:langdistro} and Table \ref{tab:multilingualstats}.

\subsection{Demographics by Country}
While our sampling strategy aimed to reflect the national ethnic distributions of each country, the limited sample size necessitated a focus on ensuring broad diversity rather than exact proportional representation. We rely on self-identified ethnicity data as provided by participants during the screening process. The distribution of ethnic groups and tribes in the study cohort is detailed in the following country-wise tables: Table \ref{tab:ghana_ethnicity} (Ghana), Table \ref{tab:kenya_ethnicity} (Kenya), Table \ref{tab:nigeria_ethnicity} (Nigeria) and Table \ref{tab:south_africa_ethnicity} (South Africa). 

\begin{center}
    \small
    \captionof{table}{Distribution of Self-Identified Ethnicity of Participants in Ghana}
    \label{tab:ghana_ethnicity}
    \begin{tabular}{|l|c|c|}
        \hline
        \textbf{Ethnicity / Tribe} & \textbf{Count} & \textbf{\%} \\
        \hline
        Akan & 31 & 30.7 \\
        Ewe & 20 & 19.8 \\
        Ga-Adangbe & 19 & 18.8 \\
        Mole-Dagbani & 17 & 16.8 \\
        Northern tribe & 3 & 3.0 \\
        Wala & 2 & 2.0 \\
        Kusasi & 2 & 2.0 \\
        Frafrah & 2 & 2.0 \\
        Kotokoli & 1 & 1.0 \\
        Kassim & 1 & 1.0 \\
        Hausa & 1 & 1.0 \\
        Guan & 1 & 1.0 \\
        Gonja & 1 & 1.0 \\
        \hline
        \textbf{Total} & \textbf{101} & \textbf{100.0} \\
        \hline
    \end{tabular}
\end{center}

\begin{center}
    \small
    \captionof{table}{Distribution of Self-Identified Ethnicity of Participants in Kenya}
    \label{tab:kenya_ethnicity}
    \begin{tabular}{|l|c|c|}
        \hline
        \textbf{Ethnicity / Tribe} & \textbf{Count} & \textbf{\%} \\
        \hline
        Kikuyu & 19 & 17.6 \\
        Kalenjin & 16 & 14.8 \\
        Luhya & 15 & 13.9 \\
        Kamba & 13 & 12.0 \\
        Mijikenda & 11 & 10.2 \\
        Luo & 11 & 10.2 \\
        Kisii & 11 & 10.2 \\
        Meru & 4 & 3.7 \\
        Maasai & 3 & 2.8 \\
        Embu & 3 & 2.8 \\
        Taita & 1 & 0.9 \\
        Arab & 1 & 0.9 \\
        \hline
        \textbf{Total} & \textbf{108} & \textbf{100.0} \\
        \hline
    \end{tabular}
\end{center}

\begin{center}
    \small
    \captionof{table}{Distribution of Self-Identified Ethnicity of Participants in Nigeria}
    \label{tab:nigeria_ethnicity}
    \begin{tabular}{|l|c|c|}
        \hline
        \textbf{Ethnicity / Tribe} & \textbf{Count} & \textbf{\%} \\
        \hline
        Hausa & 22 & 22.0 \\
        Yoruba & 20 & 20.0 \\
        Igbo & 20 & 20.0 \\
        Ijaw & 10 & 10.0 \\
        Tiv & 5 & 5.0 \\
        Kanuri & 5 & 5.0 \\
        Ibibio & 5 & 5.0 \\
        Fulani & 5 & 5.0 \\
        Urhobo & 1 & 1.0 \\
        Ogoja & 1 & 1.0 \\
        Igala & 2 & 2.0 \\
        Benin & 1 & 1.0 \\
        Nupe & 1 & 1.0 \\
        Ekoi & 1 & 1.0 \\
        Ebira & 1 & 1.0 \\
        \hline
        \textbf{Total} & \textbf{100} & \textbf{100.0} \\
        \hline
    \end{tabular}
\end{center}

\begin{center}
    \small
    \captionof{table}{Distribution of Self-Identified Ethnicity of Participants in South Africa}
    \label{tab:south_africa_ethnicity}
    \begin{tabular}{|l|c|c|}
        \hline
        \textbf{Ethnicity / Tribe} & \textbf{Count} & \textbf{\%} \\
        \hline
        Zulu & 25 & 24.8 \\
        Xhosa & 12 & 11.9 \\
        Tswana & 10 & 9.9 \\
        Afrikaans & 10 & 9.9 \\
        Pedi & 9 & 8.9 \\
        Basotho & 7 & 6.9 \\
        Tsonga & 5 & 5.0 \\
        Sesotho & 4 & 4.0 \\
        Venda & 3 & 3.0 \\
        Coloured & 3 & 3.0 \\
        Ndebele & 2 & 2.0 \\
        Koi-San & 2 & 2.0 \\
        Indian-Asian & 2 & 2.0 \\
        English & 2 & 2.0 \\
        Bantu & 2 & 2.0 \\
        Xitsonga & 1 & 1.0 \\
        Ndele & 1 & 1.0 \\
        African & 1 & 1.0 \\
        \hline
        \textbf{Total} & \textbf{101} & \textbf{100.0} \\
        \hline
    \end{tabular}
\end{center}

\section{AI Usage in Writing}
\label{app:aiusage}
We used a conversational AI tool to help lightly rephrase or shorten a few sentences, find alternate terms, and improve table formatting.

\end{appendix}

\end{document}